\documentclass[10pt,twocolumn,letterpaper]{article}

\usepackage{cvpr}              
\usepackage[accsupp]{axessibility} 
\usepackage{graphicx}
\usepackage{amsmath}
\usepackage{amssymb}
\usepackage{booktabs}
\usepackage{times}
\usepackage{epsfig}
\usepackage{graphicx}
\usepackage{mmstyle}
\usepackage{algorithm} 
\usepackage{algorithmic} 
\usepackage{multirow}

\usepackage[pagebackref,breaklinks,colorlinks]{hyperref}

\usepackage[capitalize]{cleveref}
\crefname{section}{Sec.}{Secs.}
\Crefname{section}{Section}{Sections}
\Crefname{table}{Table}{Tables}
\crefname{table}{Tab.}{Tabs.}

\def\cvprPaperID{*****} 
\def\confName{CVPR}
\def\confYear{2023}

\begin{document}

\title{Adaptive Graph Convolutional Subspace Clustering}
\author{Lai Wei, Zhengwei Chen, Jun Yin, Changming Zhu, Rigui Zhou, Jin Liu\\
	Shanghai Maritime University\\
	Haigang Avenue 1550, Shanghai, China\\
	{\tt\small weilai@shmtu.edu.cn, 965976272@qq.com, junyin@shmtu.edu.cn, cmzhu@shmtu.edu.cn}\\
    {\tt\small rgzhou@shmtu.edu.cn, jinliu@shmtu.edu.cn}
}
\maketitle

\begin{abstract}
    Spectral-type subspace clustering algorithms have shown excellent performance in many subspace clustering applications. The existing spectral-type subspace clustering algorithms either focus on designing constraints for the reconstruction coefficient matrix or feature extraction methods for finding latent features of original data samples. In this paper, inspired by graph convolutional networks, we use the graph convolution technique to develop a feature extraction method and a coefficient matrix constraint simultaneously. And the graph-convolutional operator is updated iteratively and adaptively in our proposed algorithm. Hence, we call the proposed method adaptive graph convolutional subspace clustering (AGCSC). We claim that, by using AGCSC, the aggregated feature representation of original data samples is suitable for subspace clustering, and the coefficient matrix could reveal the subspace structure of the original data set more faithfully. Finally, plenty of subspace clustering experiments prove our conclusions and show that AGCSC \footnote{We present the codes of AGCSC and the evaluated algorithms on \url{https://github.com/weilyshmtu/AGCSC}.} outperforms some related methods as well as some deep models.  %
\end{abstract}

\section{Introduction}
\label{sec:intro}

Subspace clustering has become an attractive topic in machine learning and computer vision fields due to its  success in a variety of applications, such as image processing \cite{DBLP:journals/tnn/LiLTZF22, DBLP:conf/cvpr/0002F0S0P20}, motion segmentation \cite{DBLP:journals/pami/ElhamifarV13,RN1710}, and face clustering\cite{RN2777}. The goal of subspace clustering is to arrange the high-dimensional data samples into a union of linear subspaces where they are generated \cite{art_1,RN2518,DBLP:conf/sigmod/AgrawalGGR98}. In the past decades, different types of subspace clustering algorithms have been proposed \cite{RN1940,DBLP:journals/tkde/ChuCYC09,DBLP:journals/tcsv/YiHCC18,DBLP:conf/iccv/Kanatani01,DBLP:journals/siamrev/MaYDF08}. Among them, spectral-type subspace clustering methods have shown promising performance in many real-world tasks.
\par Suppose a data matrix $\mathbf{X}=[\mathbf{x}_1;\mathbf{x}_2;\cdots;\mathbf{x}_n] \in \mathcal{R}^{n\times d}$ contains $n$ data samples drawn from $k$ subspaces, and $d$ is the number of features. The general formulation of a spectral-type subspace clustering algorithm could be expressed as follows:
\begin{equation}
    \begin{array}{cc}
        \min_{\mathbf{C}} & \Omega\big(\Phi(\mathbf{X}) - \mathbf{C}\Phi(\mathbf{X})\big)+ \lambda\Psi(\mathbf{C}), 
    \end{array}
\end{equation}\label{e1}
where $\Phi(\cdot)$ is a function that is used to find the meaningful latent features for original data samples. It could be either a linear or a non-linear feature extraction method \cite{RN1794, DBLP:conf/cvpr/YinGGHX16,DBLP:journals/tnn/XiaoTXD16}, or even a deep neural network\cite{RN2777}. $\mathbf{C}\in \mathcal{R}^{n\times n}$ is the reconstruction coefficient matrix and $\Psi(\mathbf{C})$ is usually some kind of constraint of $\mathbf{C}$. In addition, $\Omega(\cdot)$ is a function to measure the reconstruction residual, and $\lambda$ is a hyper-parameter. After $\mathbf{C}$ is obtained, an affinity graph $\mathbf{A}$ is defined as $\mathbf{A}=(|\mathbf{C}|+|\mathbf{C}^{\top}|)/2$, where $\mathbf{C}^{\top}$ is the transpose of $\mathbf{C}$. Then a certain spectral clustering, e.g. Normalized cuts (Ncuts) \cite{DBLP:journals/pami/ShiM00} is used to produce the final clustering results. 
\par Classical spectral-type subspace clustering algorithms mainly focus on designing $\Psi(\mathbf{C})$ to help $\mathbf{C}$ to carry certain characteristics and hope $\mathbf{C}$ can reveal the intrinsic structure of the original data set. For example, sparse subspace clustering (SSC) \cite{DBLP:journals/pami/ElhamifarV13} lets $\Psi(\mathbf{C}) = \|\mathbf{C}\|_1$, which makes $\mathbf{C}$ a sparse matrix. In low-rank representation (LRR) \cite{RN1710}, $\Psi(\mathbf{C})$ is the nuclear norm of $\mathbf{C}$ which helps discover the global structure of a data set. Least square regression (LSR) \cite{DBLP:conf/eccv/LuMZZHY12} aims to find a dense reconstruction coefficient matrix by setting $\Psi(\mathbf{C}) = \|\mathbf{C}\|_F^2$. Block diagonal representation (BDR) \cite{RN2485} makes $\Psi(\mathbf{C})$ a $k$-block diagonal regularizer to pursue a $k$-block diagonal coefficient matrix.
\par Recently, deep subspace clustering algorithms (DSCs) reported much better results than the classical spectral-type subspace clustering algorithms. The main difference between (DSCs) and the classical spectral-type subspace clustering algorithms is that DSCs use deep auto-encoders to extract latent features from original data \cite{RN2777, RN2739, RN2737}. But it is pointed out that the success of DSCs may be attributed to the usage of an ad-hoc post-processing strategy \cite{DBLP:conf/iclr/HaeffeleYV21}. Though the rationality of the existing DSCs is required further discussion, some deep learning techniques are still worthy of being introduced into spectral-type subspace clustering algorithms. 
\par
\begin{figure}[htbp]
    \centering
    \includegraphics[width=0.5\textwidth]{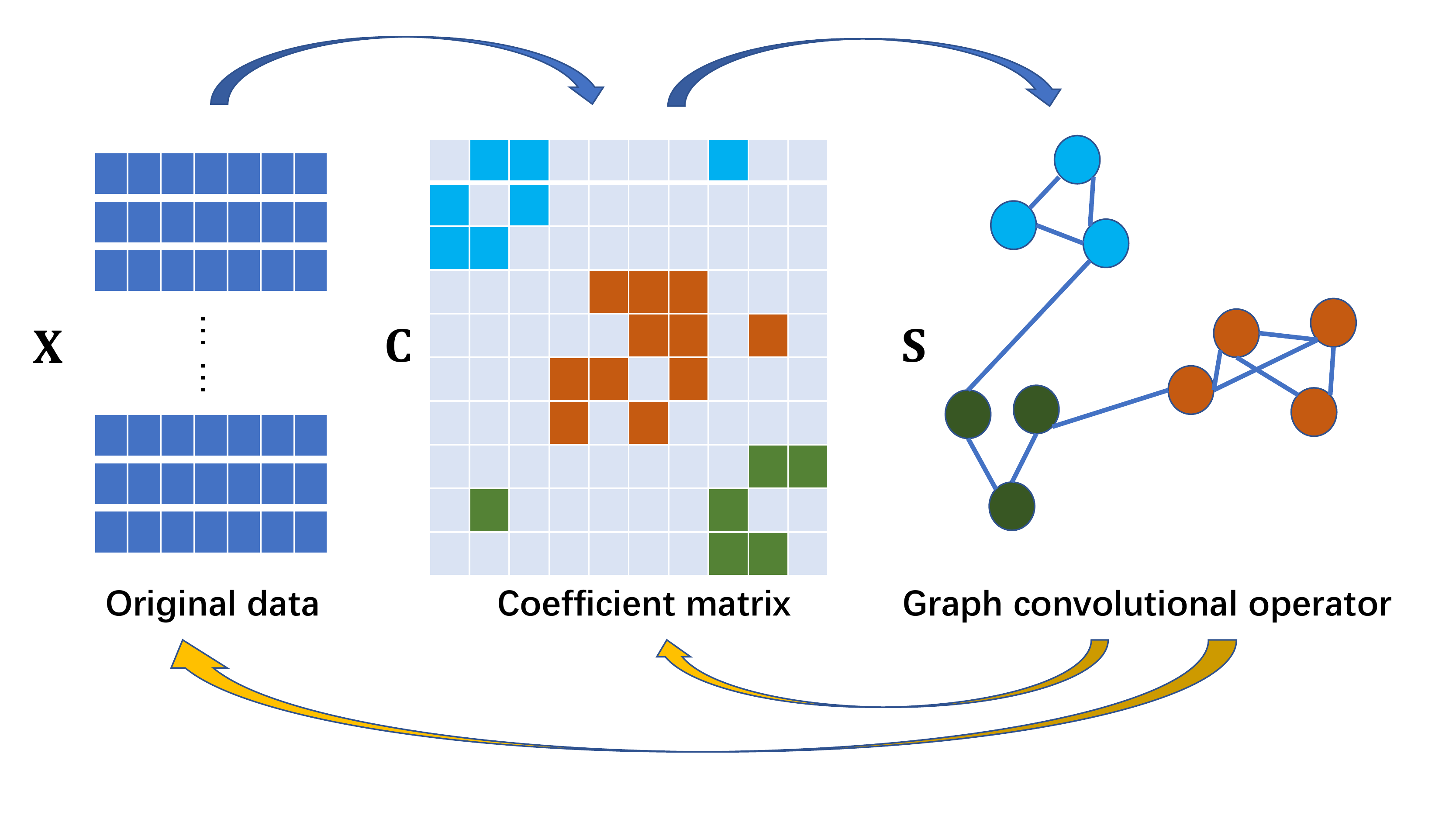}
    \caption{The overview of the proposed method. The graph convolutional operator $\mathbf{S}$ will be updated iteratively based on $\mathbf{C}$. And the updated $\mathbf{S}$ will in turn affect the computation of $\mathbf{C}$ and feature aggregation.}\label{fig1}
\end{figure}

 In this paper, inspired by graph convolutional networks \cite{DBLP:conf/iclr/KipfW17, DBLP:conf/nips/DefferrardBV16, RN2851}, we explore the problem of using graph convolutional techniques to design the feature extraction function $\Theta(\cdot)$ and the constraint function $\Psi(\cdot)$ simultaneously. Different from the existing graph convolution methods which need a predefined affinity graph, we apply the required coefficient matrix $\mathbf{C}$ to construct a graph convolutional operator. So the graph convolutional operator will be updated adaptively and iteratively in the proposed subspace clustering algorithm. Consequently, on one hand, by applying the adaptive graph convolutional operator, the aggregated feature representations of original data samples in the same subspace will be gathered closer, and those in different subspaces will be separated further. On the other hand, in the obtained coefficient matrix, the coefficients corresponding to different data samples will also have a similar characteristic to the samples' feature representations, so it could reveal the intrinsic structure of data sets more accurately. The overview pipeline of the proposed method is illustrated in Fig. \ref{fig1}.

\section{Related Work}
\label{sec:related}
\subsection{Graph convolutional networks (GCNs)}
\label{sec2.1}
GCNs \cite{DBLP:conf/iclr/KipfW17} learn new feature representations for a group of feature vectors $\mathbf{X}$ by a multi-layer network. At the $l$-th layer of a GCN model, the features $\mathbf{H}^{l-1}$ ($\mathbf{H}^{0}=\mathbf{X}$) of each node are averaged with the feature vectors in its local neighborhood first. Then the aggregated features will be transformed linearly. Finally, a nonlinear activation (e.g. ReLU) is applied to output new feature representations $\mathbf{H}^{l}$. In summary, the updating function could be expressed as follows:
\begin{equation}\label{e2}
    \mathbf{H}_{l} \leftarrow \sigma(\mathbf{S}\mathbf{H}_{l-1}\mathbf{W}_{l-1}),
\end{equation}
where $\mathbf{S} = \mathbf{\tilde{D} }^{-\frac{1}{2}}\mathbf{\tilde{A}}\mathbf{\tilde{D}}^{-\frac{1}{2}}$, $\mathbf{\tilde{A}} = \mathbf{A} + \mathbf{I}$, $\mathbf{\tilde{D}}$ is the degree matrix of $\mathbf{\tilde{A}}$. We here call $\mathbf{S}$ a \textbf{graph convolutional operator, (GCO)}. In addition, $\mathbf{A}\in \mathcal{R}^{n\times n}$ is a pre-defined adjacent matrix, $\mathbf{I}\in \mathcal{R}^{n\times n}$ is an identity matrix, $\mathbf{W}_{l-1}$ is the linear transform matrix in the $l$-th layer, $\sigma(\cdot)$ represents the nonlinear activate function. 
\par SGC (simple graph convolution) claims that ``the nonlinearity between GCN layers is not critical - but that the majority of the benefit arises from the local averaging" \cite{DBLP:conf/icml/WuSZFYW19}. Therefore, SGC simplifies a graph convolutional layer as
\begin{equation}\label{e3}
    \mathbf{H}_{l} = \mathbf{S}\mathbf{H}_{l-1}\mathbf{W}_{l-1}.
\end{equation}
By integrating several graph convolutional layers, the final feature representations is 
\begin{equation}\label{e4}
    \mathbf{F}=\overbrace{\mathbf{S}\cdots\mathbf{S}}^M\mathbf{X}\mathbf{W}_1\cdots\mathbf{W}_M = \mathbf{S}^M\mathbf{X}\mathbf{W}
\end{equation}
where $M$ is the number of graph convolutional layers, the $\mathbf{W}=\mathbf{W}_1\cdots\mathbf{W}_M$ is also a linear transformation matrix. Compared to the traditional GCNs, SGC is much simple and achieves state-of-the-art results in many real-world applications.

\subsection{Graph convolutional subspace clustering algorithms}
\label{sec2.2}
Enlightened by SGC, Cai et al. devised a graph convolutional subspace clustering algorithm (GCSC) \cite{RN2711}. In GCSC, the new feature representations are directly obtained by 1-order GCO without linear transformation. Then GCSC defines an LSR-liked model. But different from LSR, the aggregated features obtained by a graph convolutional operator is used as the basis to reconstruct the original data samples.
\par Ma et al. proposed a graph filter LSR (FLSR) algorithm \cite{DBLP:conf/mm/MaKLTC20}, which uses the coefficient matrix obtained by LSR to design the graph filter and then get the aggregated features. The two steps in FLSR are iteratively updated and the final obtained coefficient matrix will be used to get clustering results. In our opinion, both FLSR and GCSC focus on using graph convolutional techniques to devise the feature extraction function $\Phi(\cdot)$. Different from the two algorithms, our proposed method will use graph convolution techniques to design the $\Phi(\cdot)$ and $\Psi(\cdot)$ simultaneously.

\section{Methodology}
\label{sec:methodology}
\subsection{The proposed method}
\label{sec3.1}
As described in Section \ref{sec:intro}, in spectral-type subspace clustering algorithms, the obtained reconstruction coefficient matrix $\mathbf{C}$ is used to define the affinity matrix $\mathbf{A} = (|\mathbf{C}| + |\mathbf{C}^{\top}|)/2$. If we impose some additional constraints on $\mathbf{C}$, such as $\mathbf{C} = \mathbf{C}^{\top}, \mathbf{C}\geqslant \mathbf{0}$, then $\mathbf{A} = \mathbf{C}$. Moreover, if $\mathbf{C}$ also satisfies $\mathbf{C1}=\mathbf{1}$ and $diag(\mathbf{C})=\mathbf{0}$, then $\mathbf{\tilde{A}} =\mathbf{A} + \mathbf{I}=\mathbf{C} + \mathbf{I}$ and $\mathbf{\tilde{D}} = 2\mathbf{I}$. Here, $\mathbf{1}\in\mathcal{R}^{n\times 1}=[1,1,\cdots,1]^{\top}$, $diag(\mathbf{C})$ is the diagonal vector of $\mathbf{C}$. Finally, we can deduce the graph convolutional operator $\mathbf{S} = \mathbf{\tilde{D}}^{-\frac{1}{2}}\mathbf{\tilde{A}}\mathbf{\tilde{D}}^{-\frac{1}{2}} = (\mathbf{C+I})/2$. 
\par After $\mathbf{S}$ is defined, on one hand, we can get the new representation of the original data matrix as $\mathbf{F} = \mathbf{SX}=\frac{1}{2}(\mathbf{C+I})\mathbf{X}\Longrightarrow 2\mathbf{F} = (\mathbf{C+I})\mathbf{X}$. Suppose $\mathbf{C}$ is a good reconstruction coefficient matrix\footnote{Namely,  its $(i,j)$-th element is zero when $\mathbf{x}_i$ and $\mathbf{x}_j$ come from the same subspace. Otherwise, the $(i,j)$-th element is non-zero.}, then the new feature representations of samples located in the same subspace will be more similar, while the new features of samples lie in a different subspace will be more different. Then similar to GCSC, we use $\mathbf{F}$ to reconstruct $\mathbf{X}$ and hope $\mathbf{C}$ also be the reconstruction coefficient matrix, namely $\mathbf{X} = \mathbf{CF}$.
\par On the other hand, in the subspace clustering domain, the reconstruction coefficient matrix is always seen as a representation of the original data matrix \cite{conf_2, RN1710}. Therefore, we could compute the new representation of $\mathbf{C}$ as $\mathbf{SC} = \frac{1}{2}(\mathbf{C+I})\mathbf{C}=\frac{1}{2}(\mathbf{C}^2+\mathbf{C})$. By applying the same assumption in the above paragraph, $\mathbf{C}$ is hoped to faithfully reveal the subspace structure of data sets, then $\mathbf{SC}$ should have a similar characteristic to $\mathbf{C}$. Therefore, we could define a constraint of $\mathbf{C}$ as $\Psi(\mathbf{C}) =\|\mathbf{C}-\mathbf{SC}\|_F^2=\|\mathbf{C}-\frac{1}{2}(\mathbf{C}^2+\mathbf{C})\|_F^2=\|\frac{1}{2}\mathbf{C}-\frac{1}{2}\mathbf{C}^2\|_F^2=\frac{1}{4}\|\mathbf{C}-\mathbf{C}^2\|_F^2$.
\par By collecting the above definitions and through some simple deductions, we could achieve the problem under the framework of Eq. (\ref{e1}) as follows:
\begin{equation}\label{e5}
    \begin{array}{ll}
        \min_{\mathbf{F}, \mathbf{C}}& \|2\mathbf{F}-(\mathbf{C+I})\mathbf{X}\|_F^2 + \alpha\|\mathbf{X} -\mathbf{C}\mathbf{F}\|_F^2\\
        &+\beta\|\mathbf{C}-\mathbf{C}^2\|_F^2,\\
        s.t. & \mathbf{C} = \mathbf{C}^{\top}, \mathbf{C1}=\mathbf{1}, \mathbf{C}\geqslant \mathbf{0}, diag(\mathbf{C})=\mathbf{0},
    \end{array}
\end{equation}
where $\alpha$ and $\beta$ are two positive parameters. In the proposed method, the graph convolutional operator $\mathbf{S} = \frac{1}{2}(\mathbf{C}+\mathbf{I})$ is updated adaptively and iteratively which is different from GCSC, hence we term Problem (\ref{e5}) adaptive graph convolutional subspace clustering (AGCSC). 
\subsection{Optimization}
\label{sec3.2}
\subsubsection{Optimization procedure}
\label{sec3.2.1}
For the sake of solving Problem (\ref{e5}), we tranfer it into the following equivalent problem:
\begin{equation}
    \begin{array}{ll}\label{e6}
        \min_{\mathbf{F}, \mathbf{C}, \mathbf{Z}}& \|2\mathbf{F}-(\mathbf{C+I})\mathbf{X}\|_F^2 + \alpha\|\mathbf{X} -\mathbf{C}\mathbf{F}\|_F^2\\
        &+\beta\|\mathbf{C}-\mathbf{CZ}\|_F^2,\\
        s.t. & \mathbf{C} = \mathbf{Z},  \mathbf{C1}=\mathbf{1},\\
             & \mathbf{Z} = \mathbf{Z}^{\top},  \mathbf{Z}\geqslant \mathbf{0}, diag(\mathbf{Z})=\mathbf{0},
    \end{array}
\end{equation}
where $\mathbf{Z}$ is an auxiliary variable. The above problem could be solved by using ADMM (alternating direction method of multipliers method \cite{DBLP:journals/corr/LinCM10}). The augment Lagrangian function of Eq. (\ref{e6}) is:
\begin{equation}\label{e7}
\begin{array}{lll}
    \mathfrak{L} &=& \|2\mathbf{F}-(\mathbf{C+I})\mathbf{X}\|_F^2 + \alpha\|\mathbf{X} -\mathbf{C}\mathbf{F}\|_F^2 \\
    & & +\beta\|\mathbf{C}-\mathbf{CZ}\|_F^2+ \left\langle \mathbf{\Gamma}, \mathbf{C - Z}\right\rangle +  \left\langle \mathbf{\Lambda},  \mathbf{C1}-\mathbf{1}\right\rangle \\
    & & + \mu/2\big(\|\mathbf{C - Z}\|_F^2 + \|\mathbf{C1}-\mathbf{1}\|_F^2  \big)
\end{array}
\end{equation}
where $\mathbf{\Gamma,\Lambda}$ are two Lagrange multipliers, $\mu>0$ is a parameter. By minimizing $\mathfrak{L}$ with respect to the variables $\mathbf{F,C,Z}$ one at a time while fixing the others at their latest values, the variables could be optimized alternately. Suppose $t$ denotes the current iteration step, the precise updating schemes for the variables and Lagrange multipliers are described as follows:
\begin{equation}\label{e8}
\left\{
    \begin{array}{lll}
        \mathbf{C}_{t+1}&= &\big(2\mathbf{X}\mathbf{X}^{\top} +2\alpha\mathbf{F}_{t}\mathbf{X}^{\top}+2\beta(\mathbf{I}-\mathbf{Z}_{t})(\mathbf{I}-\mathbf{Z}_{t})^{\top}\\
        & &+\mu_t(\mathbf{I}+\mathbf{11}^{\top})\big)\big(4\mathbf{F}_t\mathbf{X}^{\top}-2\mathbf{X}\mathbf{X}^{\top}+2\alpha\mathbf{X}\mathbf{F}_t^{\top}\\
        & &+\mu_l(\mathbf{Z}_t+\mathbf{11}^{\top})-\mathbf{\Gamma}_t-\mathbf{\Lambda}_t\mathbf{1}^{\top}\big)^{-1},\\
        \mathbf{F}_{t+1}&=&\big(\alpha\mathbf{C}_{t+1}^{\top}\mathbf{C}_{t+1}+2\mathbf{I}\big)^{-1}\big((\mathbf{C}_{t+1} + \mathbf{I}\\
        &&+\alpha\mathbf{C}_{t+1}^{\top})\mathbf{X}\big).\\
        \mathbf{Z}_{t+1}&=&\big(2\beta\mathbf{C}_{t+1}^{\top}\mathbf{C}_{t+1}+\mu_{t}\mathbf{I}\big)^{-1}\big(2\beta\mathbf{C}_{t+1}^{\top}\mathbf{C}_{t+1}\\
        & &+\mathbf{\Gamma}_t +\mu_t\mathbf{C}_{t+1}\big).\\
        \mathbf{\Gamma}_{t+1} & = & \mathbf{\Gamma}_t + \mu_t(\mathbf{C}_{t+1}-\mathbf{Z}_{t+1}),\\
        \mathbf{\Lambda}_{t+1} & = &\mathbf{\Lambda}_t + \mu_t(\mathbf{C}_{t+1}\mathbf{1}-\mathbf{1}),\\
        \mu_{t+1} &= &\min(\mu_{max},\rho\mu_{t}),
    \end{array}
\right.
\end{equation}
where  $\mu_{max}, \rho$ are two given positive parameters. Moreover, after $\mathbf{Z}_{t+1}$ is computed, we further let $\mathbf{Z}_{t+1} = \mathbf{Z}_{t+1} - Diag(diag(\mathbf{Z}_{t+1})), \mathbf{Z}_{t+1} = \max(\mathbf{Z}_{t+1},\mathbf{0}),\mathbf{Z}_{t+1} = \big(\mathbf{Z}_{t+1} + \mathbf{Z}_{t+1}^{\top}\big)/2$, where $Diag(\cdot)$ reformulates a vector to be the diagonal of a matrix. Then $\mathbf{Z}_{t+1}$ satisfies the constraints w.r.t. $\mathbf{Z}$ in Problem (\ref{e6}).
\subsubsection{Algorithm}
\label{sec3.2.2}
\par We summarize the algorithmic procedure for solving Problem (\ref{e6}) in Algorithm \ref{algo1}. After $\mathbf{C}$ is obtained, an affinity matrix is defined as $\mathbf{A}=(|\mathbf{C}| + |\mathbf{C}^{\top}|)/2$. Then final clustering could be produced by applying Ncuts on $\mathbf{A}$.
\begin{algorithm}
    \renewcommand{\algorithmicrequire}{\textbf{Input:}}
    \renewcommand\algorithmicensure {\textbf{Output:}}
    \caption{Adaptive graph convolutional subspace clustering}\label{algo1}
    \begin{algorithmic}[1]
        \REQUIRE The data matrix $\mathbf{X}=[\mathbf{x}_1;\mathbf{x}_2;\cdots;\mathbf{x}_n]$, two parameters $\alpha,\beta>0$, the maximal number of iteration $Maxiter$;
        \ENSURE The coefficient matrix $\mathbf{C}_*$ and the new representation $\mathbf{F}_*$;
        \STATE Initialize the parameters, i.e., $t=0$, $\mu_t=10^{-6}$, $\mu_{max}=10^{30}, \rho=1.1, \varepsilon=10^{-7}$ and $\mathbf{\Gamma}_t=\mathbf{Z}_t=\mathbf{0},\mathbf{\Lambda}_t=\mathbf{0},\mathbf{F}_t=\mathbf{X}$.
        \WHILE{$\|\mathbf{C}_t-\mathbf{Z}_t\|_{\infty}>\varepsilon$,  $\|\mathbf{C}_t\mathbf{1}-\mathbf{1}\|_{\infty}>\varepsilon$ and  $t<Maxiter$}
            \STATE $t = t + 1$;
            \STATE Update variables $\mathbf{C}_t,\mathbf{F}_t,\mathbf{Z}_t$, Lagrange multipliers $\mathbf{\Gamma}_t, \mathbf{\Lambda}_t$ and parameter $\mu_t$by using Eq. (\ref{e8});
        \ENDWHILE
    \RETURN $\mathbf{C}_{*}=\mathbf{C}_{t}$ and $\mathbf{F}_* = \mathbf{F}_{t}$.
    \end{algorithmic}
    \end{algorithm}

\par We know that the convergence of ADMM for two variables has been well studied \cite{DBLP:journals/corr/LinCM10}, however there are three variables in Problem (\ref{e6}). Fortunately, we could see that all the terms in the objective function of Problem (\ref{e6}) are strongly convex. Based on the \textbf{Theorem 4.1} presented in \cite{DBLP:journals/jota/HanY12}, it could be deduced the optimization procedure (Algorithm \ref{algo1}) is convergent. 
\par Moreover, the computation time of Algorithm \ref{algo1} is mainly rely on the updating of three variables $\mathbf{C, F}$ and $\mathbf{Z}$. We can see they all have closed form solutions in Eq.(e8). For updating each variable, it takes both $O(n^3)$ to compute the pseudo-inverse of an $n \times n$ matrix and the multiplication of two $n \times n$ matrices. Hence the time complexity of Algorithm \ref{algo1} in each iteration taken together is $O(n)^3$. In our experiments, the iterations of Algorithm \ref{algo1} is always less than $500$, hence its complexity is $O(n)^3$.

\subsection{Further Disscussion}
We discuss the properties of the coefficient matrix obtained by Algorithm \ref{algo1}.
\label{sec3.3}

\subsubsection{Block diagonal property}
\par Minimizing $\Psi(\mathbf{C}) =\|\mathbf{C}-\mathbf{C}^2\|_F^2$ will lead $\mathbf{C}$ to be an approximate idempotent matrix. $\mathbf{C}=\mathbf{I}$ and $\mathbf{C}=\mathbf{0}$ are two trivial solutions to Problem (\ref{e5}). Fortunately, constraints in Problem (\ref{e5}) will prevent $\mathbf{C}$ to be the trivial solutions. Moreover, we have the following proposition, namely:
\par \textbf{Proposition 1.} \emph{The coefficient matrix $\mathbf{C}$ satisfies $\Psi(\mathbf{C})$ will be block diagonal.}
\par Firstly, in order to prove the proposition, we rewrite Problem (\ref{e5}) as follows:
\begin{equation}\label{e9}
    \begin{array}{lll}
        \min_{\mathbf{C}} & \|\mathbf{C} - \mathbf{C}^2 \|_F^2 \\
        s.t. &2\mathbf{F} = (\mathbf{C+I})\mathbf{X}, \mathbf{X} = \mathbf{CF},\\
        &\mathbf{C} = \mathbf{C}^{\top}, \mathbf{C1}=\mathbf{1}, \mathbf{C}\geqslant \mathbf{0}, diag(\mathbf{C})=\mathbf{0}.
    \end{array}
\end{equation}
By combine the two equations in the second rows of Problem (\ref{e9}), we have $\mathbf{X} = \frac{1}{2}(\mathbf{C}^2+\mathbf{C})\mathbf{X} \Rightarrow \mathbf{X} - \mathbf{CX} + \frac{1}{2}\mathbf{CX} - \frac{1}{2}\mathbf{C}^2\mathbf{X}=\mathbf{0}\Rightarrow (\frac{1}{2}\mathbf{C} + \mathbf{I})(\mathbf{X} - \mathbf{CX})=\mathbf{0}$. Because $(\frac{1}{2}\mathbf{C} + \mathbf{I})$ is invertible, hence $\mathbf{X} - \mathbf{CX}=\mathbf{0}$. Therefore, Problem (\ref{e5}) could be regarded as a relax problem of the following problem:
\begin{equation}\label{e10}
    \begin{array}{lll}
        \min_{\mathbf{C}} & \|\mathbf{C}-\mathbf{C}^2\|_F^2\\
        s.t. & \mathbf{X}=\mathbf{CX},\\
             & \mathbf{C} = \mathbf{C}^{\top}, \mathbf{C1}=\mathbf{1}, \mathbf{C}\geqslant \mathbf{0}, diag(\mathbf{C})=\mathbf{0}.
    \end{array}
\end{equation}
Problem (\ref{e10}) falls into the general subspace clustering formulation summarized in \cite{RN2485}, namely
\begin{equation}\label{e11}
    \begin{array}{cccc}
        \min_{\mathbf{C}} & f(\mathbf{X,C}), & s.t. & \mathbf{X}=\mathbf{CX}.
    \end{array}
\end{equation}
\par Secondly, for any permutation matrix $\mathbf{P}$, it is easily to verify that 
Problem (\ref{e10}) satisfies $f(\mathbf{X,C}) = f(\mathbf{PX},\mathbf{PCP}^{\top})$ which is the first EBD (enforced block diagonal) condition \cite{RN2485}.
\par Thirdly, as described in Section \ref{sec3.2.2}, Algorithm \ref{algo1} is convergent. Hence, Problem (\ref{e5}) has an unique solution $(\mathbf{F}^*, \mathbf{C}^*)$. For the same reason, Problem (\ref{e10}) also has an unique solution. 
\par According to the above explanations and \textbf{Theorem 3} in \cite{RN2485}, we can conclude that the solution to Problem (\ref{e10}) will be block diagonal. Then as a relaxed problem, solving Problem (\ref{e5}) can also get an approximate block diagonal coefficient matrix.

\subsubsection{Doubly stochastic property}
The constraints $\mathbf{C} = \mathbf{C}^{\top}, \mathbf{C1}=\mathbf{1}, \mathbf{C}\geqslant \mathbf{0}$ in Problem (\ref{e5}) restrict the reconstruction coefficient matrix $\mathbf{C}$ to be a doubly stochastic matrix. Doubly stochastic matrices have a guarantee of a certain level of connectivity, which prohibits solutions with all-zero rows or columns (that can occur in other subspace clustering methods) \cite{DBLP:journals/corr/abs-2011-14859}. And doubly stochastic normalization has been shown to significantly improve the performance of spectral clustering \cite{DBLP:conf/iccv/ZassS05,DBLP:conf/nips/ZassS06a,DBLP:conf/kdd/WangNH16}.
\par Moreover, due to the doubly stochastic property, for the $i$-th block of $\mathbf{C}$, we have $|[\mathbf{C}_i]_{p,q} - [\mathbf{C}_i]_{s,t}|\leq 1$, where $[\mathbf{C}_i]_{p,q}$ and $[\mathbf{C}_i]_{s,t}$ are the $(p,q)$-th and $(s,t)-$th elements in $\mathbf{C}_i$ and $p,q,s,t \in {1,2,\cdots, n_i}$, $n_i$ denotes the number of sample in the $i-$th subspace and $\mathbf{C}_i$ is the $i$-th block on $\mathbf{C}$. This means the differences in coefficients located in the same diagonal block will be small. This property can overcome the over-sparsity and get dense blocks \cite{RN2192} which is a benefit for obtaining good clustering results.

\subsubsection{Post-processing strategy}
As we mentioned in Section \ref{sec:intro}, post-processing could improve the performance of the existing subspace clustering algorithms \cite{conf_1, DBLP:conf/iclr/HaeffeleYV21}. The frequently used post-processing strategy is to keep the $m$-largest values for each coefficient vector and discard the relatively small ones \cite{DBLP:conf/mm/MaKLTC20, RN2281}. The rationality of the post-processing strategy is because ``the coefficients over intra-subspace data points are larger than those over inter-subspace data points.'' \cite{RN2281}. The obtained coefficient matrices of AGCSC are approximate block-diagonal, this implies that the larger coefficients are computed over intra-subspace data points and small coefficients are computed over inter-subspace data points. Hence, we also could use this kind of thresholding strategy to enhance the performance of AGCSC. In the following experiments, we call AGCSC with thresholding post-processing skill TAGCSC.

\section{Experiments}
\label{sec:experiment}
In this section, we perform subspace clustering experiments to demonstrate the effectiveness of adaptive graph convolutional subspace clustering (AGCSC).
\subsection{Dataset}
\label{sec4.1} 
We use several benchmark databases to verify the effectiveness of our proposed model. The data sets include three face image datasets (ORL, the extended Yale B and Umist), two object image datasets (COIL-20 and COIL-40), and one handwritten digit dataset (MNIST). ORL dataset \cite{orldatabase} contains 400 face images with different poses and expressions of 40 persons. The extended Yale B (YALEB) \cite{EYALEBdatabase} dataset has 2432 images of 38 persons with each person 64 near frontal images. Umist has 480 images with varied poses from 20 individuals. COIL20 dataset \cite{coildatabase} is composed of 1440 images for 20 different objects. The images of each object are taken $5^{\circ}$ apart as the object rotates on a turn table and each object has 72 images. COIL40 is similar to COIL20 which contains 2880 images for 40 different objects. MNIST dataset\footnote{\url{http://yann.lecun.com/exdb/mnist/}} comprises 10 subjects, corresponding to 10 handwritten digits, namely ``0"-``9". For this database, we use the first 100 samples of each digit in the training data sets. The detailed information on these databases is summarized in Table \ref{t1}.
\begin{table}
    \begin{center}
     \caption{Detailed information of the sevent benchmark datasets}\label{t1}   
    \begin{tabular}{cccc}
        \hline
        Dataset & Samples & Classes & Size \\ \hline
        ORL     & 400     & 40      & $32 \times 32$ \\
        YALEB  & 2432    & 38      & $48 \times 42$ \\
        Umist   & 480     & 20      & $32 \times 32$ \\
        COIL20  & 1440    & 20      & $32 \times 32$ \\
        COIL40  & 2880    & 40      & $32 \times 32$ \\
        MNIST   & 1000    & 10      & $28 \times 28$ \\
      \hline
    \end{tabular}
    \end{center}
\end{table}
\par In addition, the pixel value of each image belonging to some databases lies in $[0, 255]$. For efficient computation, we let each pixel value be divided by $255$, so that the pixel value of each image fall into $[0, 1]$. This does not change the distribution of the original data sets.

\subsection{Comparision methods and evaluation metrics}
\label{sec4.2} 
The representative and related subspace clustering models are compared in our experiments, including SSC \cite{conf_1}, LRR \cite{conf_2}, LSR \cite{DBLP:conf/eccv/LuMZZHY12}, low rank subspace clustering (LRSC) \cite{RN1940}, BDR \cite{RN2485}, thresholding ridge regression (TRR) \cite{RN2281}, TRR integrated with the graph filter method (FTRR), GCSC \cite{RN2711}, FLSR \cite{DBLP:conf/mm/MaKLTC20}. AGCSC with thresholding post-processing skill (TAGCSC) is also evaluated.
\par Two popular metrics, i.e., clustering accuracy (ACC) and normalized mutual information (NMI), are applied to quantitatively evaluate the models' performance. 
\par ACC is defined as follows:
\begin{equation}\label{e12}
    ACC =\frac{1}{n}\sum_{i=1}^n\delta(f(L_{pred}(i)),L_{true}(i)),
\end{equation}
where $L_{true}$ and $L_{pred}$ denote the ground truth labels and the prediction labels of an algorithm. 
$L_{pred}(i)$ and $L_{true}(i)$ represent $i$-th points of $L_{true}$ and $L_{pred}$ respectively. $\delta(\cdot,\cdot)$ is the indicator function that satisfies $\delta(x, y) = 1$ if $x = y$, and $\delta(x, y) = 0$ otherwise, and $f(\cdot)$ is the best mapping function that permutes clustering labels to match the ground truth labels. 
\par NMI is defined as
\begin{equation}\label{e13}
    NMI = \frac{\mathbb{I}(L_{pred}, L_{true})}{\sqrt{\mathbb{H}(L_{pred})\mathbb{H}(L_{true})}},
\end{equation}
where $\mathbb{I}(\cdot,\cdot)$ is the mutual information that measures the information gain after knowing the partitions generated by an algorithm. $\mathbb{H}(L_{pred})$ and $\mathbb{H}(L_{true})$ is the entropy of $L_{true}$ and $L_{pred}$ respectively. 
\subsection{Parameter setting}
Because parameters will influence the performance of the evaluated algorithms, hence we let the two parameters $\alpha$ and $\beta$ in AGCSC vary in the set $\{1e-5, 1e-4, 1e-3, 5e-3, 0.01, 0.05, 0.1, 0.5\}$. And for the other evaluated methods, we will tune all the parameters by following the suggestions in the corresponding references. Especially, except BDR, the other compared models all have one hyper-parameter, we let the hyper-parameters in these algorithm change in the set $\{0.001, 0.01, 0.1, 1, 5, 10, 20, 50, 80, 100\}$. For BDR, the two parameters are selected in $\{0.1,1,10,20,50, 80\}$ and $\{0.001, 0.01, 0.1,0.5,1,5,10,20,50\}$ respectively. Additionally, the neighborhood size used in GCSC is fixed as $7$. And for the thresholding skill-related methods including TRR, FTRR and TAGCSC, we let the threshold value $m$ change from $4$ to $10$. Then in different experiments, the best performance of each algorithm will be recorded. The experiments are conducted on a Windows-based machine with an Intel i9-10900 CPU, 64-GB memory and MATLAB R2021b.

\subsection{Clustering Results}
\label{sec4.3}
\begin{table*}
    \begin{center}
     \caption{Clustering results (in $\%$) of various methods on the used benchmark data sets. The best results are emphsized in bold and the second best results are denoted in bold and italic. }\label{t2} 
     \scriptsize  
    \begin{tabular}{cc|ccccccccccc}
        \hline
        \multirow{2}*{Dataset} & \multirow{2}*{Metric}  &  \multicolumn{11}{c}{Method}\\ \cline{3-13}
                            &       &   SSC     & LRR    & LRSC   &  LSR  & BDR   & TRR     & FLSR    & FTRR   & GCSC  & AGCSC   & TAGCSC \\ \hline
        \multirow{2}*{ORL}  & ACC   &   72.50     & 72.75    & 77.00   & 75.50   & 78.25   & \textbf{\emph{85.75}}     & 73.50    &   79.75   & 73.75  &  80.50   & \textbf{86.25}     \\
                            & NMI   &   84.52     & 83.26    & 85.02   & 84.97   & 88.46   & \textbf{\emph{91.49}}     & 83.84    & 87.60   & 83.98  & 88.51   & \textbf{92.84}\\ \hline
        \multirow{2}*{YALEB}& ACC   &   55.15    & 73.48    & 75.64   & 74.05   & 76.56   & 91.65    & 72.94    &  \textbf{\emph{91.90}}   & 62.50  &  84.79   & \textbf{92.31}     \\
                            & NMI   &   55.71     & 77.11    & 78.32   & 78.13   & 80.34   & 93.03     & 76.57    & \textbf{\emph{93.18}}   & 68.04  & 87.37   & \textbf{94.04}\\\hline
        \multirow{2}*{Umist}  & ACC   &   52.92    & 64.79   & 63.33   & 64.17   & 64.92   & 74.38     & 60.62    &   69.37   & 79.58  &  \textbf{\emph{81.04}}   & \textbf{90.83}     \\
                            & NMI   &   75.38     & 73.41    & 72.02   & 73.17   & 75.13   & 80.63     & 70.72    & 78.49   & 86.44  & \textbf{\emph{87.46}}   & \textbf{94.99}\\\hline

        \multirow{2}*{COIL20}  & ACC   &   68.61     & 70.14    & 71.81   & 69.17   & 71.71   & 85.97     & 69.93    &   86.53   & 79.79  &  \textbf{\emph{88.75}}   & \textbf{98.96}     \\
                            & NMI   &  66.85     & 76.43    & 77.27   & 74.17   & 80.51   & 90.23     & 77.19    & 91.17   & 85.67  & \textbf{\emph{93.38}}   & \textbf{99.11}\\\hline
        \multirow{2}*{COIL40}  & ACC   &   63.13     & 60.42    & 58.23   & 56.88   & 57.25   & 65.00     & 62.88   &   71.39   & 73.72  &  \textbf{\emph{78.12}}   & \textbf{92.60}     \\
                            & NMI   &   82.28     & 76.29    & 74.48   & 75.87   & 76.73   & 79.83     & 76.26    & 82.46   & 84.32  & \textbf{\emph{89.21}}   & \textbf{97.32}\\\hline

        \multirow{2}*{MNIST}  & ACC   &   63.70     & 64.60    & 64.30   & 62.80   & 61.30   & 67.70     & 65.10    &   66.40   & 67.70  &  \textbf{\emph{71.40}}   & \textbf{72.80}     \\
                            & NMI   &  59.75     & 60.67    & 58.91  & 57.18   & 54.76   & 64.43     & 61.10    & 63.21   & 61.99  & \textbf{\emph{65.84}}   & \textbf{67.54}\\
      \hline
    \end{tabular}
    \end{center}
\end{table*}
We first summarize the clustering results in Table \ref{t2}, where the best results are emphasized in bold and the second best results are denoted in bold and italics. From Table \ref{t2}, we can get the following observations:
\par \textbf{1.} TAGCSC constantly achieves the best results and AGCSC also outperforms the other evaluated algorithms on all the data sets except ORL and YALEB. This shows that 1) the thresholding skill could improve the performance of AGCSC; 2) and it also implies that the obtained coefficient matrices obtained by AGCSC are block-diagonal, namely, the obtained coefficients by AGCSC over intra-subspace data points are larger than those over inter-subspace data points. Moreover, if we ignore those models using the thresholding skill (TRR, FTRR), we can see AGCSC outperforms the other algorithms. This means the coefficient matrices obtained by AGCSC could faithfully reveal the subspace structures of different data sets. 
\par
\begin{figure}[htbp]
    \flushleft 
    \includegraphics[width=0.5\textwidth]{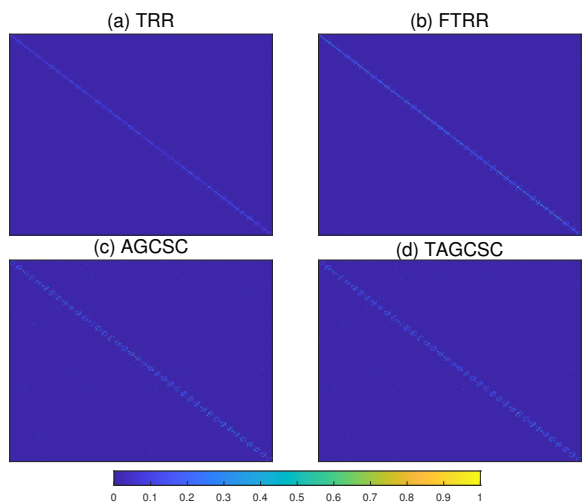}
    \caption{The obtained coefficient matrices obtained by (a) TRR, (b) FTRR, (c) AGCSC and (d) TAGCSC}\label{fig2}
\end{figure}

\begin{figure}[htbp]
    \flushleft 
    \includegraphics[width=0.5\textwidth]{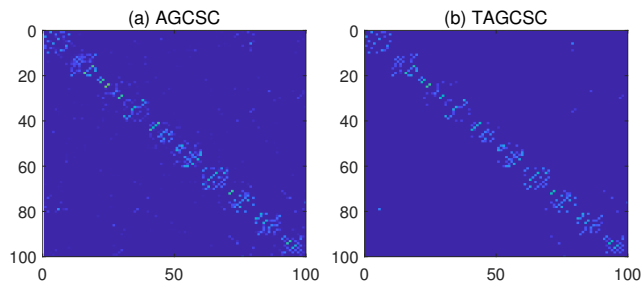}
    \caption{The partial coefficient matrices of (a) AGCSC and (b) TAGCSC corresponding to the samples from the fisrt $10$ classes.}\label{fig3}
\end{figure}

 Take ORL as an example, we show the learned reconstruction coefficient matrices obtained by AGCSC, TAGCSC, and two other algorithms that get the competitive results (TRR and FTRR) in Fig. \ref{fig2}. For a clear comparison, we use the same color bar to denote the values in the four coefficient matrices. We can see the coefficient matrix obtained by AGCSC shows a more vital block diagonal characterization. For illustrating the effect of the thresholding skill, the partial coefficient matrices of AGCSC and TAGCSC corresponding to the samples from the first $10$ classes are zoomed in Fig. \ref{fig3}. It can be seen that most coefficients that are eliminated are related to the samples existing in different subspaces.   
\par
\begin{figure}[htbp]
    \centering 
    \includegraphics[width=0.5\textwidth]{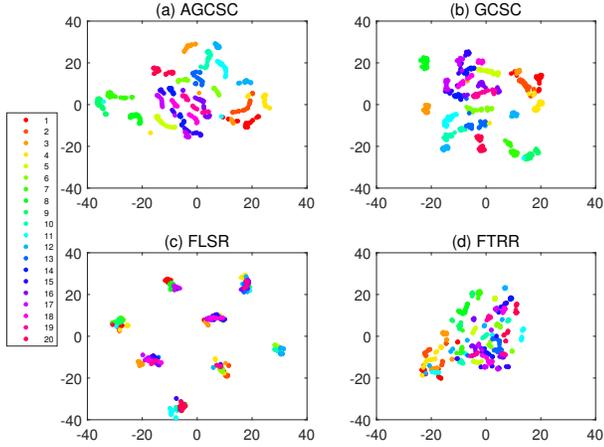}
    \caption{The visulazations of aggregated feature representations obtained by (a) AGCSC, (b) GCSC, (c) FLSR and (d) FTRR on Umist data set.}\label{fig4}
\end{figure}

 \textbf{2.} On COIL20, COIL40 and Umist data sets, AGCSC and TAGCSC get excellent results. AGCSC even dominates the models with thresholding skills such as TRR and FTRR. Moreover, on the results obtained by AGCSC, TAGCSC increases the ACC and NMI by over $10\%$ and $6\%$ respectively. Specifically, on COIL40 data set, TAGCSC outperforms all the other methods including AGCSC by a large margin, the ACC and NMI are improved by over $18.5\%$ and $9\%$ respectively. Besides the effect of thresholding skill, we believe the good performance of AGCSC and TAGCSC also relies on the aggregated feature representation computed by AGCSC. We use t-SNE \cite{JMLR:v9:vandermaaten08a} to visualize the aggregated feature representations from Umist data set obtained by AGCSC, GCSC, FLSR, and FTRR\footnote{The rest algorithms use original feature representations of samples.} respectively in Fig. \ref{fig4}. As we can see the aggregated feature representations obtained by AGCSC display clear cluster structure, while for the other algorithms, the aggregated feature representations come different subspaces are overlap.

\subsection{Parameter analysis}
\begin{figure}[htbp]
    \centering 
    \includegraphics[width=0.5\textwidth]{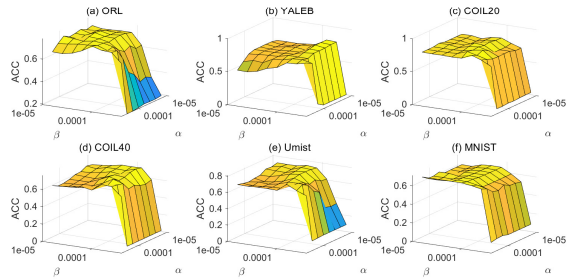}
    \caption{The influence of parameters $\alpha$ and $\beta$ on clustering accuracy of AGCSC.}\label{fig5}
\end{figure}

\begin{figure}[htbp]
    \centering 
    \includegraphics[width=0.5\textwidth]{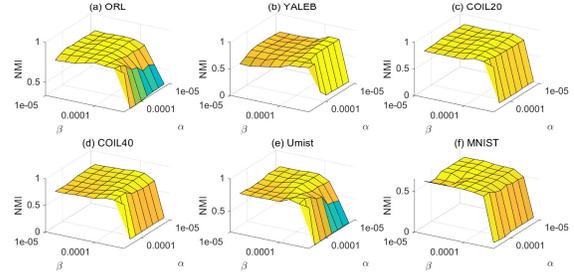}
    \caption{The influence of parameters $\alpha$ and $\beta$ on normalized multual information of AGCSC.}\label{fig6}
\end{figure}

AGCSC has two parameters $\alpha$ and $\beta$. We show the effects of $\alpha$ and $\beta$ on the clustering performance of AGCSC. Fig. \ref{fig5} and Fig. \ref{fig6} show the clustering accuracies and normalized mutual information obtained by AGCSC varied with the parameters respectively. It is clearly that on all the data sets, AGCSC achieves better results when $\alpha$ and $\beta$ take relatively small values.
\par
\begin{figure}[htbp]
    \centering 
    \includegraphics[width=0.5\textwidth]{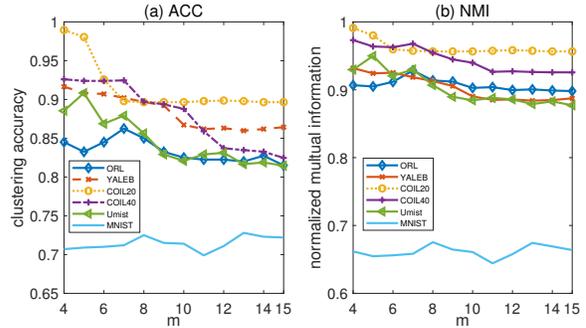}
    \caption{The influence of thresholding value $m$ on the clustering performance of TAGCSC.}\label{fig7}
\end{figure}

 TAGCSC has an additional parameter, i.e., the thresholding value $m$. For a certain value of $m$, we record clustering results of TAGCSC on each pair of $(\alpha, \beta)$. Then the best result could be determined for a fixed $m$. Then we let $m$ vary from 4 to 15 and illustrate the clustering perfomance of TAGCSC in Fig. \ref{fig7}. It can be found that TAGCSC get better results when $m$ is relatively small on all data sets except MNIST. And the performance of TAGCSC is much stable on MNIST dataset. Based on these results, we suggest $m$ to be a relative small value (i.e., $m\leq 8$) for subspace clustering tasks.

\subsection{Convergence and complexity analysis}
\begin{figure}[htbp]
    \centering 
    \includegraphics[width=0.3\textwidth]{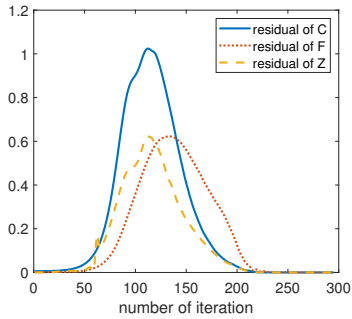}
    \caption{The residuals of variables $\mathbf{F,C,Z}$ versus the iterations on ORL database.}\label{fig8}
\end{figure}

Firstly, we discuss the convergence of AGCSC. Take ORL as an example, we record the computed variables $\mathbf{F,C,Z}$ of Algorithm \ref{algo1} in each iteration. Then the residuals of three variables (i.e., $\|\mathbf{F}_{t+1}-\mathbf{F}_t\|_F^2$,$\|\mathbf{C}_{t+1}-\mathbf{C}_t\|_F^2$ and $\|\mathbf{Z}_{t+1}-\mathbf{Z}_t\|_F^2$) could be obtained. The residuals versus the number of iterations are illustrated in Fig. \ref{fig8}. We can see that after about 250 iterations, Algorithm \ref{algo1} converges into a stable solution.
\par
\begin{figure}[htbp]
    \centering 
    \includegraphics[width=0.5\textwidth]{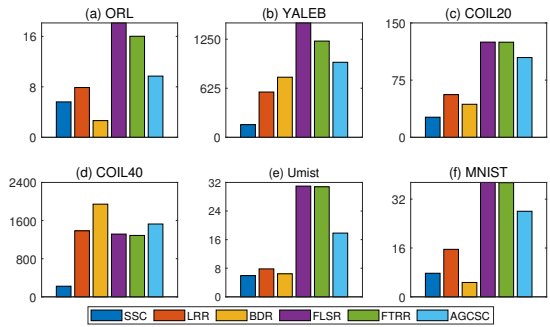}
    \caption{Computation time (seconds) for different algorithms to run on different datasets. The y-axis coordinates are in seconds.}\label{fig9}
\end{figure}

 Secondly, we compare the complexity of AGCSC with the other evaluated algorithms. Because LSR, LRSC, and TRR have closed-form solutions, we do not consider the consumption time of these algorithms. Then the average computation time (seconds) of the rest algorithms to run on different datasets is shown in Fig. \ref{fig9}. We can see that 1) the computation time of AGCSC is less than those of FLSR and FTRR on all the data sets except COIL40; 2) the complexity of AGCSC is in the same order as the time complexity of the other evaluated algorithms.

\subsection{Comparison with deep clustering methods}
\begin{table*}
    \begin{center}
        \caption{Clustering results (in $\%$) of AGCSC and TAGCSC compared with several deep clustering methods. The best results are emphsized in bold. Some results are ignored because the results are not found in corresponding literatures.}\label{t3}
        \scriptsize
        \begin{tabular}{cc|ccccccccc} 
\hline
\multirow{2}{*}{Dataset}&  \multirow{2}{*}{Metric} & \multicolumn{7}{c}{Method} \\ \cline{3-11}
    &                   &         AE+SSC & DSC-L1  & DSC-L2  &  DEC    & DKM    & DCCM   & PSSC & AGCSC   & TAGCSC \\\hline
\multirow{2}{*}{ORL}    & ACC  & 75.63   & 85.50   & 86.00   &  51.75  & 46.82  & 62.50  & \textbf{86.75} & 80.50   & 86.25\\
                        & NMI  & 85.55   & 90.23   & 90.34   &  74.49  & 73.32  & 79.06  & \textbf{93.49} & 88.51   & 92.84\\\hline
\multirow{2}{*}{YALEB}  & ACC  & 74.80   & 96.80   & \textbf{97.33}   &  86.84  & -    & -   & -& 84.79   & 92.32\\
                        & NMI  & 78.33   & 96.87   & \textbf{97.03}   &  92.40  & -    & -   & -& 87.34   & 94.04\\\hline
\multirow{2}{*}{Umist}  & ACC  & 70.42   & 72.42   & 73.12   &  55.21  & 51.06  & 54.48 & 79.17  & 81.04   & \textbf{90.83}\\
                        & NMI  & 75.15   & 75.56   & 76.62   &  71.25  & 72.49  & 74.40 & 86.70  & 87.49   & \textbf{94.99}\\\hline
\multirow{2}{*}{COIL20} & ACC  & 87.11  & 93.14   & 93.68   &  72.15  & 66.51  & 80.21  & 97.22  & 88.75   & \textbf{98.96}\\
                        & NMI  & 89.90   & 93.53   & 94.08   &  80.07  & 79.71  & 86.39 & 97.79  & 93.38   & \textbf{99.11}\\\hline
\multirow{2}{*}{COIL40} & ACC  & 73.91  & 80.03   & 80.75   &  48.72  & 58.12  & 76.91  & 83.58  & 78.12   & \textbf{92.60}\\
                        & NMI  & 83.18  & 88.52   & 89.41   &  74.17  & 78.40  & 88.90  & 92.58  & 89.21   & \textbf{97.23}\\\hline
\multirow{2}{*}{MNIST}  & ACC  & 48.40  & 72.80   & 75.00   &  61.20  & 53.32  & 40.20  & \textbf{84.30} & 71.40   & 72.80\\
                        & NMI  & 53.37  & 72.17   & 73.19   &  57.43  & 50.02  & 34.68  & \textbf{76.76} & 65.84   & 67.54\\\hline
        \end{tabular}
    \end{center}
\end{table*}
Due to the promising results achieved by deep clustering models, we here compare AGCSC and TAGCSC with some deep clustering algorithms including SSC with pre-trained convolutional auto-encoder
features (AE+SSC), deep subspace clustering network (DSC) \cite{DBLP:conf/nips/JiZLSR17}, deep embedding clustering (DEC) \cite{DBLP:conf/icml/XieGF16}, deep $K$-means (DKM) \cite{DBLP:journals/prl/FardTG20}, deep comprehensive correlation mining (DCCM) \cite{DBLP:conf/iccv/WuLWQLLZ19} and pseudo-supervised deep subspace clustering (PSSC) \cite{DBLP:journals/tip/LvKLX21}. The experimental results of the deep models are cited from some latest literatures \cite{DBLP:conf/nips/JiZLSR17,RN2737,RN2856,DBLP:conf/mm/MaKLTC20}. Table \ref{t3} shows the comparison results.
\par We can observer from Table \ref{t3} that 1) TAGCSC outperforms the deep models on Umist, COIL20 and COIL40 datasets, and is close to DSC on MNIST datasets. This shows the effectiveness of TAGCSC. Considering the complexity and the unexplained characteristics of DNNs, our proposed algorithm is much appealing; 2) TAGCSC is inferior to DSC-L1 and DSC-L2 on YALEB dataset. However, as pointed in \cite{DBLP:conf/iclr/HaeffeleYV21}, the success of DSC-L1 and DSC-L2 may be attributed to to post-processing method. Without post-processing, DSC can only achieve $59.09\%$ clustering accuracy on YALEB. This result is much worse than those of AGCSC and TAGCSC. Hence, we can say that our proposed algorithm has a comparable performance to those of the deep models.
\section{Conclusions}
In this paper, we develop a new subspace clustering algorithm, named adaptive graph convolutional subspace clustering (AGCSC). In AGCSC, we propose to use the reconstruction coefficient matrix to design a graph convolutional operator. Then use the graph convolutional operator to smooth the feature representations and the coefficient matrix simultaneously. We demonstrate plenty of subspace clustering experiments to show 
the superiorities of feature representations and coefficient matrix obtained by AGCSC. And we claim that AGCSC and its extension (TAGCSC) are comparable to several deep clustering models. 
\section{Acknowledge}
This work is supported by Shanghai Municipal Natural Science Foundation (20ZR1423100), National Key Research and Development Program of China (2021YFC2801000) and Shanghai Pujiang Program (Grant No. 22PJD029).

{\small
\bibliographystyle{ieee_fullname}
\bibliography{CameraReady}

\begin{thebibliography}{10}\itemsep=-1pt

\bibitem{DBLP:conf/sigmod/AgrawalGGR98}
Rakesh Agrawal, Johannes Gehrke, Dimitrios Gunopulos, and Prabhakar Raghavan.
\newblock Automatic subspace clustering of high dimensional data for data
  mining applications.
\newblock In {\em Proceedings {ACM} {SIGMOD} International Conference on
  Management of Data, June 2-4, 1998, Seattle, Washington, {USA}}, pages
  94--105. {ACM} Press, 1998.

\bibitem{RN2711}
Yaoming Cai, Zijia Zhang, Zhihua Cai, Xiaobo Liu, Xinwei Jiang, and Qin Yan.
\newblock Graph convolutional subspace clustering: A robust subspace clustering
  framework for hyperspectral image.
\newblock {\em IEEE Transactions on Geoscience and Remote Sensing}, pages
  1--12, 2020.

\bibitem{DBLP:journals/tkde/ChuCYC09}
Yi{-}Hong Chu, Yi{-}Ju Chen, De{-}Nian Yang, and Ming{-}Syan Chen.
\newblock Reducing redundancy in subspace clustering.
\newblock {\em {IEEE} Trans. Knowl. Data Eng.}, 21(10):1432--1446, 2009.

\bibitem{DBLP:conf/nips/DefferrardBV16}
Micha{\"{e}}l Defferrard, Xavier Bresson, and Pierre Vandergheynst.
\newblock Convolutional neural networks on graphs with fast localized spectral
  filtering.
\newblock In {\em Annual Conference on Neural Information Processing Systems
  2016, December 5-10, 2016, Barcelona, Spain}, pages 3837--3845, 2016.

\bibitem{conf_1}
Ehsan Elhamifar and Ren{\'{e}} Vidal.
\newblock Sparse subspace clustering.
\newblock In {\em Proc. {IEEE} Computer Society Conference on Computer Vision
  and Pattern Recognition ({CVPR} 2009)}, pages 2790--2797, Miami, Florida,
  USA, June 2009.

\bibitem{DBLP:journals/pami/ElhamifarV13}
Ehsan Elhamifar and Ren{\'{e}} Vidal.
\newblock Sparse subspace clustering: algorithm, theory, and applications.
\newblock {\em IEEE Trans Pattern Anal Mach Intell}, 35(11):2765--81, 2013.

\bibitem{DBLP:journals/prl/FardTG20}
Maziar~Moradi Fard, Thibaut Thonet, and {\'{E}}ric Gaussier.
\newblock Deep \emph{k}-means: Jointly clustering with \emph{k}-means and
  learning representations.
\newblock {\em Pattern Recognit. Lett.}, 138:185--192, 2020.

\bibitem{DBLP:conf/iclr/HaeffeleYV21}
Benjamin~David Haeffele, Chong You, and Ren{\'{e}} Vidal.
\newblock A critique of self-expressive deep subspace clustering.
\newblock In {\em 9th International Conference on Learning Representations,
  {ICLR} 2021, Virtual Event, Austria, May 3-7, 2021}. OpenReview.net, 2021.

\bibitem{DBLP:journals/jota/HanY12}
Deren Han and Xiaoming Yuan.
\newblock A note on the alternating direction method of multipliers.
\newblock {\em J. Optim. Theory Appl.}, 155(1):227--238, 2012.

\bibitem{DBLP:conf/nips/JiZLSR17}
Pan Ji, Tong Zhang, Hongdong Li, Mathieu Salzmann, and Ian~D. Reid.
\newblock Deep subspace clustering networks.
\newblock In {\em Annual Conference on Neural Information Processing Systems
  2017, December 4-9, 2017, Long Beach, CA, {USA}}, pages 24--33, 2017.

\bibitem{DBLP:conf/iccv/Kanatani01}
Ken{-}ichi Kanatani.
\newblock Motion segmentation by subspace separation and model selection.
\newblock In {\em Proceedings of the Eighth International Conference On
  Computer Vision (ICCV-01), Vancouver, British Columbia, Canada, July 7-14,
  2001 - Volume 2}, pages 586--591. {IEEE} Computer Society, 2001.

\bibitem{DBLP:conf/iclr/KipfW17}
Thomas~N. Kipf and Max Welling.
\newblock Semi-supervised classification with graph convolutional networks.
\newblock In {\em {ICLR} 2017, Toulon, France, April 24-26, 2017, Conference
  Track Proceedings}. OpenReview.net, 2017.

\bibitem{EYALEBdatabase}
Kuang{-}Chih Lee, Jeffrey Ho, and David~J. Kriegman.
\newblock Acquiring linear subspaces for face recognition under variable
  lighting.
\newblock {\em {IEEE} Trans. Pattern Anal. Mach. Intell.}, 27(5):684--698,
  2005.

\bibitem{DBLP:journals/tnn/LiLTZF22}
Jun Li, Hongfu Liu, Zhiqiang Tao, Handong Zhao, and Yun Fu.
\newblock Learnable subspace clustering.
\newblock {\em {IEEE} Trans. Neural Networks Learn. Syst.}, 33(3):1119--1133,
  2022.

\bibitem{DBLP:journals/corr/abs-2011-14859}
Derek Lim, Ren{\'{e}} Vidal, and Benjamin~D. Haeffele.
\newblock Doubly stochastic subspace clustering.
\newblock {\em CoRR}, abs/2011.14859, 2020.

\bibitem{DBLP:journals/corr/LinCM10}
Zhouchen Lin, Minming Chen, and Yi Ma.
\newblock The augmented lagrange multiplier method for exact recovery of
  corrupted low-rank matrices.
\newblock {\em CoRR}, abs/1009.5055, 2010.

\bibitem{RN1710}
Guangcan Liu, Zhouchen Lin, Shuicheng Yan, Ju Sun, Yong Yu, and Yi Ma.
\newblock Robust recovery of subspace structures by low-rank representation.
\newblock {\em IEEE Transaction on Pattern Analysis and Machine Intelligence},
  35(1):171--184, 2013.

\bibitem{conf_2}
Guangcan Liu, Zhouchen Lin, and Yong Yu.
\newblock Robust subspace segmentation by low-rank representation.
\newblock In {\em (ICML-10), June 21-24, 2010,}, pages 663--670, Haifa, Israel,
  jun 2010.

\bibitem{RN2485}
Canyi Lu, Jiashi Feng, Zhouchen Lin, Tao Mei, and Shuicheng Yan.
\newblock Subspace clustering by block diagonal representation.
\newblock {\em IEEE Transactions on Pattern Analysis and Machine Intelligence},
  41(2):487--501, 2019.

\bibitem{DBLP:conf/eccv/LuMZZHY12}
Can{-}Yi Lu, Hai Min, Zhong{-}Qiu Zhao, Lin Zhu, De{-}Shuang Huang, and
  Shuicheng Yan.
\newblock Robust and efficient subspace segmentation via least squares
  regression.
\newblock In {\em Computer Vision - {ECCV} 2012 - 12th European Conference on
  Computer Vision, Florence, Italy, October 7-13, 2012, Proceedings, Part
  {VII}}, volume 7578, pages 347--360. Springer, 2012.

\bibitem{RN2737}
J. Lv, Z. Kang, X. Lu, and Z. Xu.
\newblock Pseudo-supervised deep subspace clustering.
\newblock {\em IEEE Trans Image Process}, 30:5252--5263, 2021.

\bibitem{DBLP:journals/tip/LvKLX21}
Juncheng Lv, Zhao Kang, Xiao Lu, and Zenglin Xu.
\newblock Pseudo-supervised deep subspace clustering.
\newblock {\em {IEEE} Trans. Image Process.}, 30:5252--5263, 2021.

\bibitem{DBLP:journals/siamrev/MaYDF08}
Yi Ma, Allen~Y. Yang, Harm Derksen, and Robert~M. Fossum.
\newblock Estimation of subspace arrangements with applications in modeling and
  segmenting mixed data.
\newblock {\em {SIAM} Review}, 50(3):413--458, 2008.

\bibitem{DBLP:conf/mm/MaKLTC20}
Zhengrui Ma, Zhao Kang, Guangchun Luo, Ling Tian, and Wenyu Chen.
\newblock Towards clustering-friendly representations: Subspace clustering via
  graph filtering.
\newblock In {\em {MM} '20, Virtual Event / Seattle, WA, USA, October 12-16,
  2020}, pages 3081--3089. {ACM}, 2020.

\bibitem{art_1}
Lance Parsons, Ehtesham Haque, and Huan Liu.
\newblock Subspace clustering for high dimensional data: A review.
\newblock {\em SIGKDD Explor. Newsl.}, 6(1):90--105, 2004.

\bibitem{RN1794}
Vishal~M. Patel, Hien~Van Nguyen, and Rene Vidal.
\newblock Latent space sparse subspace clustering.
\newblock In {\em ICCV}, pages 225--232, 2014.

\bibitem{RN2777}
X. Peng, J. Feng, J.~T. Zhou, Y. Lei, and S. Yan.
\newblock Deep subspace clustering.
\newblock {\em IEEE Trans Neural Netw Learn Syst}, 31(12):5509--5521, 2020.

\bibitem{RN2281}
Xi Peng, Zhang Yi, and Huajin Tang.
\newblock Robust subspace clustering via thresholding ridge regression.
\newblock {\em AAAI}, pages 3827--3833, 2015.

\bibitem{RN2739}
Zhihao Peng, Yuheng Jia, Hui Liu, Junhui Hou, and Qingfu Zhang.
\newblock Maximum entropy subspace clustering network.
\newblock {\em IEEE Transactions on Circuits and Systems for Video Technology},
  pages 1--1, 2021.

\bibitem{RN2856}
Z. Peng, H. Liu, Y. Jia, and J. Hou.
\newblock Adaptive attribute and structure subspace clustering network.
\newblock {\em IEEE Trans Image Process}, 31:3430--3439, 2022.

\bibitem{coildatabase}
H.~Murase S.A.~Nene, S.K.~Nayar.
\newblock Columbia object image library (coil-20).
\newblock {\em Technical Report CUCS}, 1996.

\bibitem{orldatabase}
F.~S. Samaria and A.~C. Harter.
\newblock Parameterisation of a stochastic model for human face identification.
\newblock In {\em Applications of Computer Vision, 1994., Proceedings of the
  Second IEEE Workshop on}, 1994.

\bibitem{DBLP:journals/pami/ShiM00}
Jianbo Shi and Jitendra Malik.
\newblock Normalized cuts and image segmentation.
\newblock {\em {IEEE} Trans. Pattern Anal. Mach. Intell.}, 22(8):888--905,
  2000.

\bibitem{RN2192}
K. Tang, D.~B. Dunson, Z. Su, R. Liu, J. Zhang, and J. Dong.
\newblock Subspace segmentation by dense block and sparse representation.
\newblock {\em Neural Network}, 75:66--76, 2016.

\bibitem{JMLR:v9:vandermaaten08a}
Laurens van~der Maaten and Geoffrey Hinton.
\newblock Visualizing data using t-sne.
\newblock {\em Journal of Machine Learning Research}, 9(86):2579--2605, 2008.

\bibitem{RN2518}
Rene Vidal.
\newblock Subspace clustering.
\newblock {\em IEEE Signal Processing Magazine}, 28(2):52--68, 2011.

\bibitem{RN1940}
Ren{\'{e}} Vidal and Paolo Favaro.
\newblock Low rank subspace clustering (lrsc).
\newblock {\em Pattern Recognition Letters}, 43:47--61, 2014.

\bibitem{DBLP:conf/kdd/WangNH16}
Xiaoqian Wang, Feiping Nie, and Heng Huang.
\newblock Structured doubly stochastic matrix for graph based clustering:
  Structured doubly stochastic matrix.
\newblock In Balaji Krishnapuram, Mohak Shah, Alexander~J. Smola, Charu~C.
  Aggarwal, Dou Shen, and Rajeev Rastogi, editors, {\em Proceedings of the 22nd
  {ACM} {SIGKDD}, San Francisco, CA, USA, August 13-17, 2016}, pages
  1245--1254. {ACM}, 2016.

\bibitem{DBLP:conf/icml/WuSZFYW19}
Felix Wu, Amauri H.~Souza Jr., Tianyi Zhang, Christopher Fifty, Tao Yu, and
  Kilian~Q. Weinberger.
\newblock Simplifying graph convolutional networks.
\newblock In {\em {ICML} 2019, 9-15 June 2019, Long Beach, California, {USA}},
  volume~97 of {\em Proceedings of Machine Learning Research}, pages
  6861--6871, 2019.

\bibitem{RN2851}
Felix Wu, Amauri Holanda~de Souza, Tianyi Zhang, Christopher Fifty, Tao Yu, and
  Kilian~Q. Weinberger.
\newblock Simplifying graph convolutional networks.
\newblock {\em ICML}, 2019.

\bibitem{DBLP:conf/iccv/WuLWQLLZ19}
Jianlong Wu, Keyu Long, Fei Wang, Chen Qian, Cheng Li, Zhouchen Lin, and
  Hongbin Zha.
\newblock Deep comprehensive correlation mining for image clustering.
\newblock In {\em 2019 {IEEE/CVF} International Conference on Computer Vision,
  {ICCV} 2019, Seoul, Korea (South), October 27 - November 2, 2019}, pages
  8149--8158. {IEEE}, 2019.

\bibitem{DBLP:journals/tnn/XiaoTXD16}
Shijie Xiao, Mingkui Tan, Dong Xu, and Zhao~Yang Dong.
\newblock Robust kernel low-rank representation.
\newblock {\em {IEEE} Trans. Neural Networks Learn. Syst.}, 27(11):2268--2281,
  2016.

\bibitem{DBLP:conf/icml/XieGF16}
Junyuan Xie, Ross~B. Girshick, and Ali Farhadi.
\newblock Unsupervised deep embedding for clustering analysis.
\newblock In Maria{-}Florina Balcan and Kilian~Q. Weinberger, editors, {\em
  Proceedings of the 33nd International Conference on Machine Learning, {ICML}
  2016, New York City, NY, USA, June 19-24, 2016}, volume~48 of {\em {JMLR}
  Workshop and Conference Proceedings}, pages 478--487. JMLR.org, 2016.

\bibitem{DBLP:journals/tcsv/YiHCC18}
Shuangyan Yi, Zhenyu He, Yiu{-}Ming Cheung, and Wen{-}Sheng Chen.
\newblock Unified sparse subspace learning via self-contained regression.
\newblock {\em {IEEE} Trans. Circuits Syst. Video Technol.}, 28(10):2537--2550,
  2018.

\bibitem{DBLP:conf/cvpr/YinGGHX16}
Ming Yin, Yi Guo, Junbin Gao, Zhaoshui He, and Shengli Xie.
\newblock Kernel sparse subspace clustering on symmetric positive definite
  manifolds.
\newblock In {\em {CVPR} 2016, Las Vegas, NV, USA, June 27-30, 2016}, pages
  5157--5164. {IEEE} Computer Society, 2016.

\bibitem{DBLP:conf/iccv/ZassS05}
Ron Zass and Amnon Shashua.
\newblock A unifying approach to hard and probabilistic clustering.
\newblock In {\em {(ICCV} 2005), 17-20 October 2005, Beijing, China}, pages
  294--301. {IEEE} Computer Society, 2005.

\bibitem{DBLP:conf/nips/ZassS06a}
Ron Zass and Amnon Shashua.
\newblock Doubly stochastic normalization for spectral clustering.
\newblock In {\em Advances in Neural Information Processing Systems 19,
  Vancouver, British Columbia, Canada, December 4-7, 2006}, pages 1569--1576.
  {MIT} Press, 2006.

\bibitem{DBLP:conf/cvpr/0002F0S0P20}
Tao Zhou, Huazhu Fu, Chen Gong, Jianbing Shen, Ling Shao, and Fatih Porikli.
\newblock Multi-mutual consistency induced transfer subspace learning for human
  motion segmentation.
\newblock In {\em {CVPR} 2020, Seattle, WA, USA, June 13-19, 2020}, pages
  10274--10283. Computer Vision Foundation / {IEEE}, 2020.

\end{thebibliography}
}

\end{document}